%% file: main.tex
\documentclass{article}
\usepackage{ijcai22}

\usepackage{times}
\usepackage{soul}
\usepackage{url}
\usepackage{hyperref}
\hypersetup{hidelinks}
\usepackage[utf8]{inputenc}
\usepackage[small]{caption}
\usepackage{graphicx}
\usepackage{amsmath}
\usepackage{amsthm}
\usepackage{booktabs}
\usepackage{algorithm}
\usepackage{algorithmic}
\usepackage{amsfonts}
\usepackage{mathtools}
\usepackage[title]{appendix}
\graphicspath{{./images/}}

\newtheorem{definition}{Definition}
\newtheorem{hypothesis}{Hypothesis}

\title{With Greater Distance Comes Worse Performance: On the Perspective of Layer Utilization and Model Generalization}

\author{  { }
James Wang$^1$
\And
Cheng-Lin Yang$^1$
\affiliations
$^1$CyCraft AI Lab, Taipei, Taiwan\\
\emails
\{james.wang, cl.yang\}@cycraft.com
}

\def\textsupsub#1#2{\rlap{\textsuperscript{#1}}\textsubscript{#2}}

\begin{document}

\maketitle

\begin{abstract}
  Generalization of deep neural networks remains one of the main open problems in machine learning. Previous theoretical works focused on deriving tight bounds of model complexity, while empirical works revealed that neural networks exhibit double descent with respect to both training sample counts and the neural network size. In this paper, we empirically examined how different layers of neural networks contribute differently to the model; we found that early layers generally learn representations relevant to performance on both training data and testing data. Contrarily, deeper layers only minimize training risks and fail to generalize well with testing or mislabeled data. We further illustrate the distance of trained weights to its initial value of final layers has high correlation to generalization errors and can serve as an indicator of an overfit of model. Moreover, we show evidence to support post-training regularization by re-initializing weights of final layers. Our findings provide an efficient method to estimate the generalization capability of neural networks, and the insight of those quantitative results may inspire derivation to better generalization bounds that take the internal structure of neural networks into consideration.
  
\end{abstract}

\input{intro}

\input{related}
\input{terminology}
\input{results}
\input{discussion}
\input{appendix}

\bibliographystyle{named}
\bibliography{citations}

\end{document}

%% file: intro.tex
\section{Introduction}

Modern deep neural networks are extremely over-parameterized and thus capable of memorizing training samples instead of extracting generalizable features. Under the traditional {\em bias-variance trade-off} theory, complex networks should easily overfit training data. However, in recent years, widespread adoption and success of neural networks in image classification, natural language processing and several other fields have proved deep networks capable of learning generalizable features, which disagrees with VC-dimension based explanations of model complexity.

Theoretical works also proposed the PAC framework \cite{valiant1984theory} to explain the concept of learnability. The subsequent research \cite{10.1145/279943.279989} constructed PAC-Bayes theorems, which allowed works such as \cite{DBLP:journals/corr/DziugaiteR17,Jiang2020Fantastic} to derive tighter bounds for generalization of neural networks using different priors. While those bounds are often magnitudes better within controlled environment, they are still loose or vacuous in many cases \cite{DBLP:journals/corr/abs-2002-08791}.

Apart from theoretical derivations, \cite{belkin2019reconciling,Nakkiran2020Deep} have discussed the ``double descent" phenomenon, where for both data complexity and the neural network capacity. The U-shaped-like risk curve only appears in the under-parameterized regime, and the testing risk constantly improves after crossing the {\em interpolation threshold}. Combining this with the monotonically decreasing training risk, we can find that generalization error peaks at the interpolation threshold, and the error decreases as models move away from it.

Meanwhile, another line of empirical researches focuses on the layer-wise view of models. \cite{zhang2019layers} showed that different layers of a neural network have different importance to the entire model. \cite{baldock2021deep} attempted to explain model behaviour with classification depth of individual data samples. At the same time, \cite{stephenson2021on} leveraged replica-based mean field theory manifold analysis to evaluate the differences between layers and explain generalization.

Main contributions of this paper are listed as follows:
\begin{itemize}
\item We illustrate that the utilization (distance of trained weights to their initial values) for deeper layers predicts the amount of non-generalizable information learnt by models with high confidence. Thus, it can be used to predict whether a model overfits.

\item We provide evidence to support post-training regularlization by reinitializing part of the weights of a trained neural network.

\item We demonstrate that deep neural networks follows a top-down order during the learning process, and refrain from using deeper layers when early layers are already capable of handling the provided training dataset.

\item The definition of model resilience is formalized as an additional method to measure the ability of a model to extract general patterns within training data.

\end{itemize}

%% file: related.tex
\section{Related Works}

Generalization of deep neural networks has been discussed in several prior research works. \cite{DBLP:conf/iclr/ZhangBHRV17} showed that neural networks are often capable of memorizing randomized data and suggested traditional methods are not capable of explaining a neural network's ability to generalize. Among research on neural network behaviours, \cite{belkin2019reconciling} were the first to study and coin the name double decent. Following their results, \cite{Nakkiran2020Deep} conducted thorough experiments on double decent for neural networks across several different settings, and proposed the ``effective model complexity" theory to explain their results. Aside from double descent, \cite{zhang2019layers} found that different layers of neural networks tend to behave differently, but did not give further explanations for their results. Recent works including \cite{baldock2021deep,stephenson2021on} started exploring the relationship between generalization and behaviour of individual layers, but only focused on the dynamics of the learning procedure. Our work extends this line of research and studies relationship between layer characteristics and generalization. Different from previous works, which monitor fluctuation of generalization error throughout the training procedure (epoch-wise double descent), we focus on comparing models, trained till convergence, with different neural network sizes and data complexity.{\em}

%% file: terminology.tex
\section{Terminology and Hypothesis}


\subsection{Definitions}

\begin{definition}[Model] \em
In this paper, the term {\em model} is strictly used to describe the combination of a hypothesis function class $\mathbb{F}$ (it is equivalent to the neural network architecture in our case), $n$ training samples (\textbf{\textit{S\textsubscript{X}}},\ \textbf{\textit{S\textsubscript{Y}}}) where $\textbf{\textit{S\textsubscript{X}}} = \lbrace x_i \rbrace^n_{i=1}$ and $\textbf{\textit{S\textsubscript{Y}}} = \lbrace y_i \rbrace^n_{i=1}$, an optimizer \textbf{\textit{O}}, and a loss function \textbf{\textit{j}}($\hat{y}$,\ $y$) where $\hat{y}$ is predicted label and $y$ is target label. Additionally, we define $\mathbb{F}$\textsupsub{$\prime$}{} as the final function selected from hypothesis class $\mathbb{F}$ after training, and \textbf{\textit{J}}(\textbf{\textit{f}}; \textbf{\textit{S\textsubscript{X}}},\textbf{\textit{S\textsubscript{Y}}}) as the average loss over (\textbf{\textit{S\textsubscript{X}}},\ \textbf{\textit{S\textsubscript{Y}}}) with respect to function \textbf{\textit{f}}.
\begin{align}
    \textbf{\textit{J}}(\textbf{\textit{f}}; \textbf{\textit{S\textsubscript{X}}},\textbf{\textit{S\textsubscript{Y}}}) = \frac{1}{n} \sum_{i=1}^{n}\ \textbf{\textit{j}}(\ \textbf{\textit{f}}(x_i),y_i)
\end{align}

\end{definition}

\begin{definition}[Neural Network Capacity] \em
We define the neural network capacity as the VC-dimension of the network. This definition captures the upper bound of a neural network's {\em ability} to fit training samples while only considering the neural network size.
\end{definition}

\begin{definition}[Data Complexity] \em
The complexity of data \textbf{\textit{C\textsubscript{S}}} is defined as the entropy \textbf{\textit{H}}(\textbf{\textit{S\textsubscript{X}, S\textsubscript{Y}}}), which can be decomposed as follows.
\begin{align}
    \textbf{\textit{H}}(\textbf{\textit{S\textsubscript{X}, S\textsubscript{Y}}}) &= \textbf{\textit{H}}(\textbf{\textit{S\textsubscript{X}}}) + \textbf{\textit{H}}(\textbf{\textit{S\textsubscript{Y}}} \lvert \textbf{\textit{S\textsubscript{X}}})
\end{align}
\textbf{\textit{H}}(\textbf{\textit{S\textsubscript{X}}}) describes the ``{\em amount}" of samples $n$ in  \textbf{\textit{S\textsubscript{X}}}, and \textbf{\textit{H}}(\textbf{\textit{S\textsubscript{Y}}}$\lvert$\textbf{\textit{S\textsubscript{X}}}) characterizes the ``{\em perplexity}" of samples, which is the difficulty of extracting patterns from \textbf{\textit{S\textsubscript{X}}} to decide corresponding labels \textbf{\textit{S\textsubscript{Y}}}.\footnote{A more formal definition for \textbf{\textit{C\textsubscript{S}}} is included in the technical appendix.}
\end{definition}

\begin{definition}[Effective Model Complexity] \em
The effective complexity of a model with respect to the neural network $\mathbb{F}$, loss function \textbf{\textit{J}}, optimizer \textbf{\textit{O}}, tolerance $\epsilon$, and a fixed training epoch $e$ is illustrated as follows.
\begin{align}
    EMC_{\mathbb{F},\textbf{\textit{J}},\textit{\textbf{O}}, \epsilon, e}
    \coloneqq \max\ \lbrace \textbf{\textit{C\textsubscript{S}}}\ \lvert\ \mathbb{E}_{\textbf{\textit{S}}\sim\textbf{\textit{D}}^n}\lbrack\  \textbf{\textit{J}}(\mathbb{F'}; \textbf{\textit{S\textsubscript{X}}},\textbf{\textit{S\textsubscript{Y}}})\rbrack\ \leq \epsilon \rbrace
\end{align}
where $\mathbb{F'}$ is the model {\em learnt} from $\mathbb{F}$.

Our definition is similar to the one proposed by \cite{Nakkiran2020Deep}, but refined in two parts. First, training procedure \textbf{\textit{$\tau$}} in the original paper is split into \textbf{\textit{J}}, \textbf{\textit{O}}, $e$, and specifically $e$ is decoupled from the rest of the components. The reason behind separating $e$ from the rest of the training procedure is that the fluctuation of generalization error during training is shown to be largely affected by the ordering of features in \cite{stephenson2021epochwise}, and \cite{chen2021multiple} proved it is possible to handcraft samples that exhibit epoch-wise multiple descent behaviour for linear models. We drew the conclusion that epoch-wise and model-wise double descent must have different root causes, and only focused on model-wise double descent in this work. Thus, it is required to choose a large enough $e$, and only measure EMC after the model is trained till convergence.

The second difference is we adopted data complexity \textbf{\textit{C\textsubscript{S}}} instead of number of training samples $n$. This is necessary for our experiment setup, and will be shown in section 4.
\end{definition}

\begin{definition}[Layer Contribution] \em
Contribution to the layer is defined as the difference between measurement \textbf{\textit{M}} on $\mathbb{F'}$ and $\mathbb{F'}$\textbf{\textsupsub{0}{L}}, where $\mathbb{F'}$\textbf{\textsupsub{0}{L}} is the model with weights of layer \textbf{L} reinitialized to its initial values. \textbf{\textit{M}} can be any metric that takes a function \textbf{\textit{f}} and any amount of additional arguments.
\begin{align}
    \textbf{\textit{M}}(\mathbb{F'}\textbf{\textsupsub{0}{L}};\  \textbf{\textit{*args}}) - \textbf{\textit{M}}(\mathbb{F'};\ \textbf{\textit{*args}})
\end{align}

For instance, taking the average loss \textbf{\textit{J}} over (\textbf{\textit{S\textsubscript{X}}},\ \textbf{\textit{S\textsubscript{Y}}}) as \textit{\textbf{M}} can be written as follow.
\begin{align}
    \textbf{\textit{J}}(\mathbb{F'}\textbf{\textsupsub{0}{L}};\  \textbf{\textit{S\textsubscript{X}}},\textbf{\textit{S\textsubscript{Y}}}) - \textbf{\textit{J}}(\mathbb{F'};\ \textbf{\textit{S\textsubscript{X}}},\textbf{\textit{S\textsubscript{Y}}})
\end{align}

Layer contribution measures how the training of a specific layer affects performance of a model on chosen metric \textbf{\textit{M}}.
\end{definition}

\begin{definition}[Layer Utilization] \em
Utilization of a layer is defined as the $l_2$-distance between trained weights \textbf{\textit{W}}\textbf{\textsupsub{$\prime$}{L}} and its initial weights \textbf{\textit{W}}\textbf{\textsupsub{0}{L}}.
\begin{align}
    \lVert\ \textbf{\textit{W}}\textbf{\textsupsub{$\prime$}{L}}-\textbf{\textit{W}}\textbf{\textsupsub{0}{L}}\ \rVert\textsubscript{2}
\end{align}

This definition is equivalent to treating the initial weights of a neural network as a prior, and measuring how much a model strays from the prior after training.
\end{definition}

\begin{definition}[Generalization] \em
Generalization is defined as the gap between the inference results on training and testing data.
\begin{align}
    \lvert\ \textbf{\textit{J}}(\mathbb{F}\textsupsub{$\prime$}{}\ ;\  \textbf{\textit{S\textsubscript{X\textsubscript{train}}}},\textbf{\textit{S\textsubscript{Y\textsubscript{train}}}}) - \textbf{\textit{J}}(\mathbb{F}\textsupsub{$\prime$}{}\ ;\  \textbf{\textit{S\textsubscript{X\textsubscript{test}}}},\textbf{\textit{S\textsubscript{Y\textsubscript{test}}}})\ \rvert
\end{align}

\end{definition}
\begin{definition}[Resilience] \em
Resilience is the ability to cast doubt on mislabeled training samples, and recognize the ``{\em correct}" labels instead.

To measure the resilience, we have to train on data with a certain proportion of intentionally incorrect (corrupted) labels, and test on the correct (recovered) labels after training ends.
\begin{align}
    \textit{\textbf{J}}(\mathbb{F}\textsupsub{$\prime$}{}\ ;\  \textbf{\textit{S\textsubscript{X\textsubscript{recovered}}}},\textbf{\textit{S\textsubscript{Y\textsubscript{recovered}}}})
\end{align}

\end{definition}

The difference between generalization and resilience is discussed in appendix A.

\subsection{Hypothesis}

\begin{hypothesis}[Monotonic Contribution] \em
For fully connected neural networks without skip layers, the contribution of neural networks layers follows a monotonic pattern, where early layers (closer to input) start taking effect early on, and deeper layers only start contributing when trained on data with higher complexity.
\end{hypothesis}

\begin{hypothesis}[Final Layer Utilization as EMC] \em
Based on Hypothesis 1, we deduced that the utilization of the final layer in a neural network should provide an estimation for EMC.
\end{hypothesis}

%% file: results.tex
\section{Results}

\begin{figure*}[!ht]
    \centering
    \includegraphics[width=480pt, height=!]{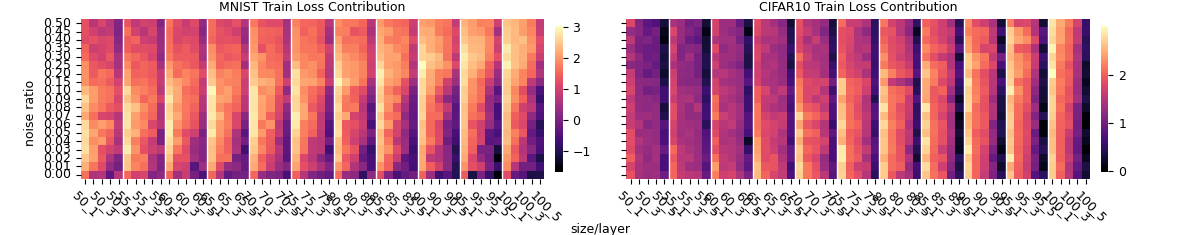}
    \caption{{\em Contribution of each model and layer to training loss.} Each block (separated by vertical white lines) represents the contribution of separate layers for different model sizes. Y-axis ticks show the noise ratio (0.5 means 50\% of training labels are shuffled), and x-axis ticks are composed of two fields separated by an underscore. The first field represents the hidden layer size, and the second field indicates the depth of the layer (50\_1 corresponds to the first layer in a model with the hidden layer size 50). Ladder-like patterns emerge on the bottom right of each block, where the contribution of deeper layers gradually grows as the noise ratio in the training data increases. Log scale is taken to allow the difference between smaller values to be visible.}
    \label{fig1}
\end{figure*}

\begin{figure*}[!ht]
    \centering
    \includegraphics[width=\textwidth,height=!]{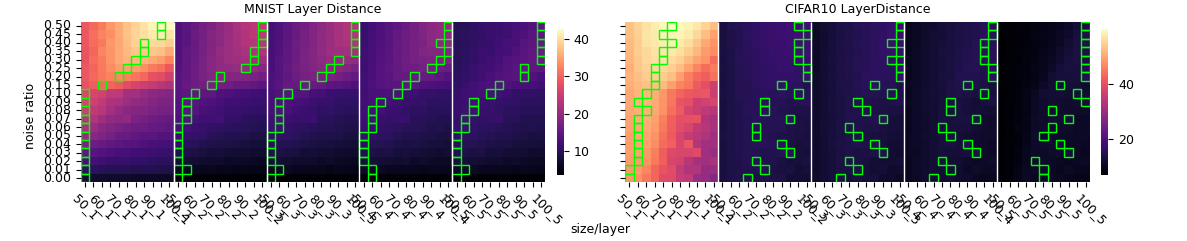}
    \caption{{\em Utilization of each model and layer.} Heatmaps are organized to display the layers of the same depth from different models in the same block (separated by vertical white lines). Cells with the green square in each block represent that they have the largest value among their rows. It clearly shows that models with the largest utilization (cells marked with green squares) form a diagonal pattern in each block, and the patterns in deeper layers tend to appear closer to the bottom right of the heatmap.}
    \label{fig2}
\end{figure*}

\subsection{Experiment Setup}

Our network is composed of 5 fully connected layers with ReLU as the activation function. The network was trained on MNIST and CIFAR10 and optimized with SGD. The main reason MNIST and CIFAR10 were chosen is that both datasets have significantly lower proportion of noisy labels than other datasets based on the findings of \cite{northcutt2021pervasive}. 

The decision of using fully connected layers instead of convolution layers is to avoid introducing additional inductive bias. As we are uncertain how those assumptions of data geometry might further introduce complexity to analysis of model behaviours. 

The neural network capacity is tuned by changing the size of hidden layers, and the data complexity is probed by shuffling a portion of training labels. The fraction of labels shuffled is called ``{\em noise ratio}". The range of hidden layer sizes and noise ratio is carefully chosen to be within our computation power, while still capable of demonstrating the transition from the under-parameterized to the over-parameterized regime. These choices are optimized for MNIST; thus, applying the same range of hidden layer size and noise ratio to CIFAR10, a more complex dataset, is expected to have more models falling into the under-parameterized regime. This tendency can be observed from figures in later sections.

Finally, each experiment was executed 3 times on MNIST and 6 times on CIFAR10, and all models were trained for 10000 epochs before performing inference.

\subsection{Layer Inequality}

Deeper layers only start contributing when the data complexity increases, which results in the pattern shown in Figure \ref{fig1}. This result indicates that neural networks with relatively high capacity refrain from storing learnt information in deeper layers, and supports the aforementioned Hypothesis 1.

Meanwhile, Figure \ref{fig2} reveals a clear pattern in utilization. Starting from the top left of each heatmap, where the under-parameterized regime is located, the utilization starts to increase until hitting the boundary marked by green squares (max utilization). Then, the utilization decreases as it moves toward the bottom right, where the over-parameterized regime is located. This pattern is further discussed in section 4.4.

\subsection{Generalization and Resilience}

\begin{figure*}[!ht]
    \centering
    \includegraphics[width=480pt, height=!]{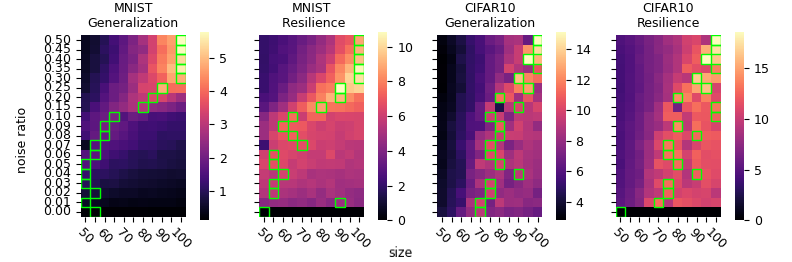}
    \caption{{\em Generalization and resilience of trained models.} The diagonal pattern from the bottom-left to the top-right indicated the existence of the interpolation threshold. Models on the left of the threshold fall in under-parameterized regime, while the right hand side belongs to the over-parameterized regime.}
    \label{fig3}
\end{figure*}

In Figure \ref{fig3}, the contribution of generalization and resilience are shown side by side to demonstrate that both measurements follow a similar pattern, where the occurrence of max value for each noise ratio forms a diagonal line. This line marks the existence of the double descent interpolation threshold.

The first observation we found from Figure \ref{fig3} is that the interpolation threshold forms a positively correlated trace between different noise ratios and neural network sizes. This trace agrees with results shown in \cite{Nakkiran2020Deep}, in which data complexity is tuned by controlling the amount of samples provided to train the model. Our experiments show that probing the noise ratio can affect trained models in a similar way to their results. This justifies the use of data complexity \textbf{\textit{C\textsubscript{S}}} instead of training samples $n$ in definition of EMC.

An important finding is also discovered by comparing Figure \ref{fig2} and \ref{fig3}. The heatmaps of both utilization and generalization/resilience also follow a similar pattern, which we will shortly discuss in section 4.4.

In addition to aforementioned observations, a noticeable effect is also worth being addressed. While both generalization and resilience possess the same diagonal pattern, the bottom right half of the resilience plot, which represents a large hidden layer and low noise ratio, is significantly brighter. This effect can be explained by exploring the training procedure. In practice, the resilience is almost always contradictory to the objective function, where the training risk must be minimized empirically. When the EMC is high enough, the model will manage to fit data with shuffled labels to achieve low training risk. The forced memorization is then reflected as having relatively worse resilience when compared to under-parameterized models. At the same time, for samples that have its training labels shuffled, over-parameterized models are generally capable of ``{\em recognizing}" correct labels (labels before shuffling) while catering to the objective function, allowing them to retain better resilience when compared to models near the interpolation threshold.

\subsection{Utilization to Generalization and Resilience}

\begin{figure*}[!ht]
    \centering
    \includegraphics[width=480pt, height=!]{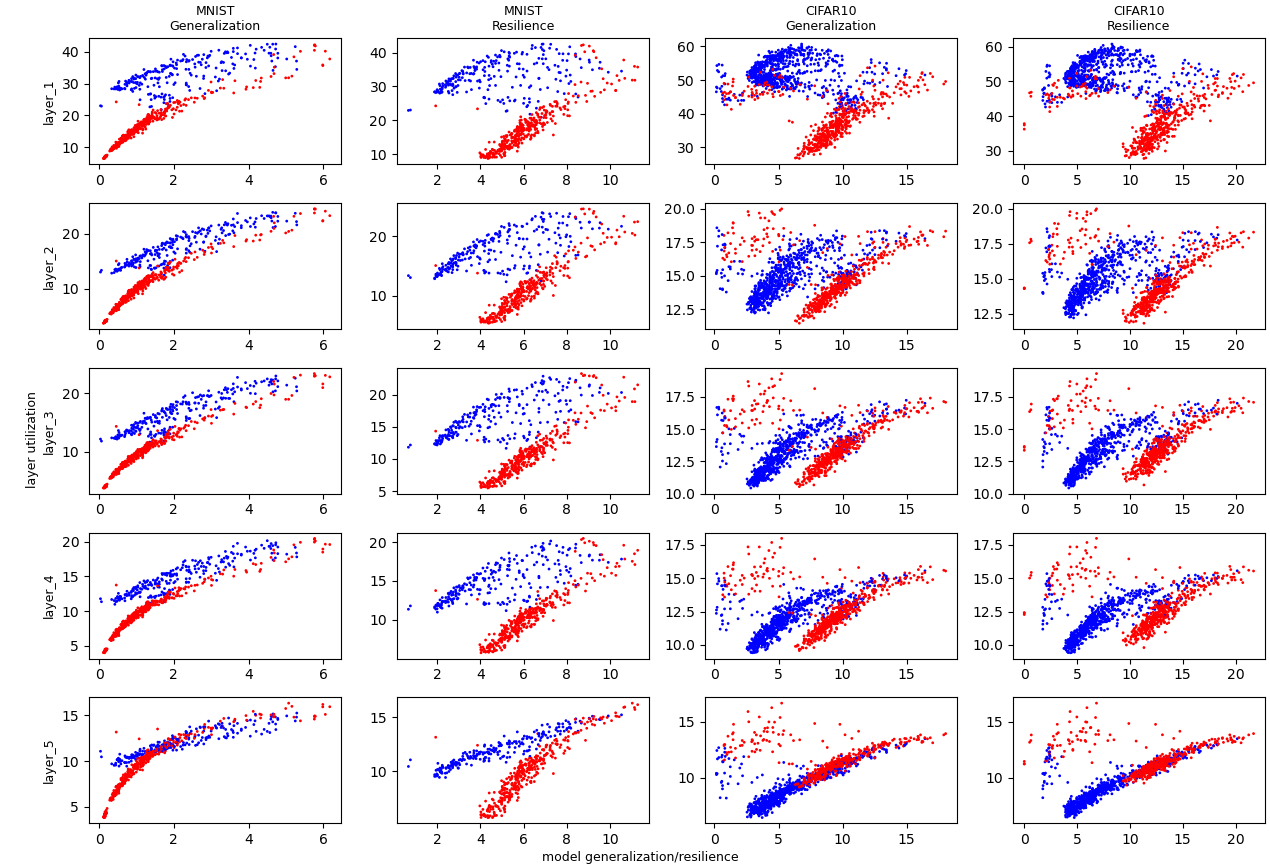}
    \caption{{\em Utilization vs. Generalization/Resilience.} The x-axis represents the measurement of generalization/resilience over MNIST/CIFAR10, and the y-axis shows the utilization of each layer. The blue points correspond to models lying on the top-left (under-parameterized) side of interpolation threshold  in Figure \ref{fig3}; red points are those on the bottom-right (over-parameterized) side.}
    \label{fig4}
\end{figure*}

Figure \ref{fig4} compares utilization to generalization and resilience. Initially, we found that models lying on different sides of the interpolation threshold converge into two different sets of patterns in early layers, which illustrates that under- and over-parameterized regimes exhibit different characteristics. As we investigated toward deeper layers, the tail of two curves gradually merged together, which is the interpolation threshold between them. The utilization of those merged curves are positively correlated to both generalization and resilience. This finding agrees with the proposed Hypothesis 2, where utilization of deeper layers estimates EMC. Therefore, it can predict the effective complexity of a model relatively well.

\subsection{Contribution and Overfit}
\begin{figure*}[!ht]
    \centering
    \includegraphics[width=480pt, height=!]{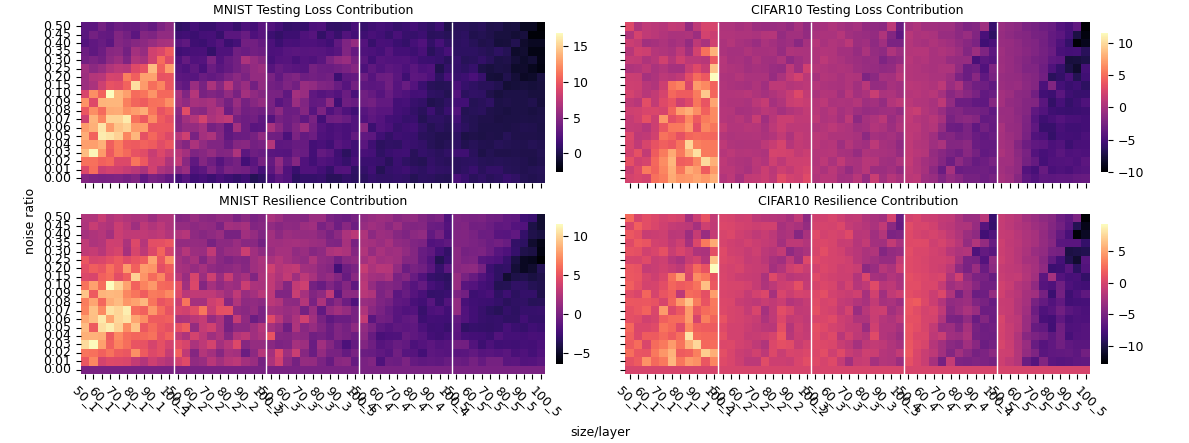}
    \caption{{\em Contribution to Testing Loss and Resilience.} The overall contribution in early layers tend to be positive while those of the deeper layers are negative. This shows that latter layers participate heavily in storing non-generalizable information, and its harm towards testing results outweighs the impact of helpful features learnt.}
    \label{fig5}
\end{figure*}

As shown in Figure \ref{fig5}, while earlier layers positively contribute to the overall result, deeper layers usually have an overall negative contribution to testing loss and resilience. The result provides insight about tendencies of trained models, where early layers tend to focus on widely applicable patterns, and deeper layers often capture the remaining {\em noise} in training data.

The observed pattern can be used as an useful indicator that allows users to gauge and revert overfit of model by simply re-initializing the last few layers. Re-initialization of deeper layers can be thought of as forcing a trained model to forget memorized noises and provide regularization to a model after the training phase.

%% file: discussion.tex
\section{Discussion}

In section 4, we have shown that layer utilization depends on the complexity of training samples and deeper layers only get activated when early layers are no longer capable of handling provided training data. Apart from learning orders, it is also shown that the utilization of final layers can predict generalization.

Combining aforementioned observations, we concluded that the utilization of deeper layers are a suitable estimation of EMC. It can be easily calculated, which makes it extremely useful in practice. Finally, we also provide the empirical proof that performing rolling back on final layers does in fact allow models to revert overfit behaviour. This agrees with the results of \cite{baldock2021deep,stephenson2021on}. While experiment results are promising, there are still several questions that are yet to be addressed.

\begin{itemize}

\item {\bf Why does the diagonal utilization pattern appear in each layer respectively, and appear to be not synchronized?}

While all layers are trained together, similar patterns appearing in all layers seem to suggest that each layer learns semi-independent representations from the output of previous layers, and there might not exist complex cross layer relations during optimization.

\item {\bf Why does re-initializing the layer weight result in good results?}

From the perspective of PAC-Bayes theories, our results suggest that the initialization of the deeper layers in a model can be treated as a prior to the distribution of hypothesis class. 

While our results can be supported by PAC-Bayes framework, it still doesn't explain why the initialization of deeper layers needs to be respected, whereas early layers are allowed to explore the parameter space further from initial weights.

Furthermore, it is still an open problem on why the trained weights of early layers can cooperate well with the initial weights of deeper layers, and retain high performance after re-initialization.

\item {\bf Are the results universally applicable for other network structures?}

All experiments in this paper were conducted on fully connected layers. It is worthwhile to explore whether other network structures, such as convolution or recurrent layers, behave the same as this paper's results.


\end{itemize}

To answer above questions, we encourage readers to explore the two domains below:

\begin{itemize}
\item {\bf The landscape illustrated by deep neural networks}

A comprehensive understanding would help understand why the parameters do not stray far from initialization in the case where a 0 training risk solution clearly exists, but EMC is below data complexity. 

\item {\bf How do redundant parameters in deep neural network help in learning?}

We hypothesize that the redundant parameters and their initialization values help in constructing a prior that makes learning ``{\em easy}" for over-parameterized settings. Additionally, the deviation from initialization in deeper layers has more critical impacts on the generalization. This hypothesis is similar to the lottery hypothesis proposed in \cite{frankle2018the}. However, more detailed research needs to be conducted to support this proposition.

\end{itemize}

We believe our work has established an empirical basis to help developing future works in this field, and hopefully inspire more research to comprehend the generalization of deep neural networks.

%% file: appendix.tex
\begin{appendices}
\section{Generalization and Resilience}
While using generalization and resilience metrics together is capable of providing a reliable view over how well a model learns generalizable features, the shortcomings and uncertainties of using generalization and resilience as empirical measurements will be discussed here. 

\subsection{Shortcomings in Empirical Measurements of Generalization and Resilience}
Measuring generalization of models requires the comparison between training risks and testing risks. However, \cite{power2022grokking} has shown the convergence of loss on training and testing data can be highly asynchronous. To the best of our knowledge, it is currently impossible to identify whether a model has fully converged with respect to testing loss.

This causes a problem when the performance of a trained model over testing data needs to be measured. It is impossible to determine whether the results have converged to a final stable state or still at an intermediate stage due to training dynamics. Measurements of generalization suffer from this uncertainty.

On the other hand, resilience is more robust in general. Since it does not require an independent set of testing samples, but is calculated on the training samples. Thus, resilience stabilizes along with training loss, and can be measured once training converges.

\subsection{Resilience and Double Descent}
While double descent exists in the measurement of resilience, the second descent is significantly more obscure compared to the measurement of generalization. To understand the cause of this effect, it is required to think about the differences between generalization within the under- and over-parameterized regime.

In the under-parameterized regime, solutions found by the model can not achieve close-to-perfect results on the training data. This implies that the model will constantly focus on general patterns rather than outliers (shuffled labels). The learnt general patterns are helpful in achieving lower risks on recovered labels, and resulting in better resilience.

In contrast, close-to-perfect solutions {\em can} be found by the model in the over-parameterized regime. This implies the risk for data with shuffled labels are also minimized, and the risk of recovered labels will be high.

However, compared to critically-parameterized models which have few good solutions to pick from, and must sacrificed learning general patterns for lower training risk, the over-parameterized models have more ``{\em luxury}" in exploring solutions that preserve general patterns and fit noises well. This allows over-parameterized models to still performing slightly better than models near the interpolation threshold.
\end{appendices}

%% file: main.bbl
\begin{thebibliography}{}

\bibitem[\protect\citeauthoryear{Baldock \bgroup \em et al.\egroup
  }{2021}]{baldock2021deep}
Robert John~Nicholas Baldock, Hartmut Maennel, and Behnam Neyshabur.
\newblock Deep learning through the lens of example difficulty.
\newblock In A.~Beygelzimer, Y.~Dauphin, P.~Liang, and J.~Wortman Vaughan,
  editors, {\em Advances in Neural Information Processing Systems}, 2021.

\bibitem[\protect\citeauthoryear{Belkin \bgroup \em et al.\egroup
  }{2019}]{belkin2019reconciling}
Mikhail Belkin, Daniel Hsu, Siyuan Ma, and Soumik Mandal.
\newblock Reconciling modern machine-learning practice and the classical
  bias--variance trade-off.
\newblock {\em Proceedings of the National Academy of Sciences},
  116(32):15849--15854, 2019.

\bibitem[\protect\citeauthoryear{Chen \bgroup \em et al.\egroup
  }{2021}]{chen2021multiple}
Lin Chen, Yifei Min, Misha Belkin, and amin karbasi.
\newblock Multiple descent: Design your own generalization curve.
\newblock In A.~Beygelzimer, Y.~Dauphin, P.~Liang, and J.~Wortman Vaughan,
  editors, {\em Advances in Neural Information Processing Systems}, 2021.

\bibitem[\protect\citeauthoryear{Dziugaite and
  Roy}{2017}]{DBLP:journals/corr/DziugaiteR17}
Gintare~Karolina Dziugaite and Daniel~M. Roy.
\newblock Computing nonvacuous generalization bounds for deep (stochastic)
  neural networks with many more parameters than training data.
\newblock In {\em Proceedings of the Thirty-Third Conference on Uncertainty in
  Artificial Intelligence}. {AUAI} Press, 2017.

\bibitem[\protect\citeauthoryear{Frankle and Carbin}{2019}]{frankle2018the}
Jonathan Frankle and Michael Carbin.
\newblock The lottery ticket hypothesis: Finding sparse, trainable neural
  networks.
\newblock In {\em International Conference on Learning Representations}, 2019.

\bibitem[\protect\citeauthoryear{Jiang \bgroup \em et al.\egroup
  }{2020}]{Jiang2020Fantastic}
Yiding Jiang, Behnam Neyshabur, Hossein Mobahi, Dilip Krishnan, and Samy
  Bengio.
\newblock Fantastic generalization measures and where to find them.
\newblock In {\em International Conference on Learning Representations}, 2020.

\bibitem[\protect\citeauthoryear{McAllester}{1998}]{10.1145/279943.279989}
David~A. McAllester.
\newblock Some pac-bayesian theorems.
\newblock In {\em Proceedings of the Eleventh Annual Conference on
  Computational Learning Theory}, COLT' 98, page 230–234, New York, NY, USA,
  1998. Association for Computing Machinery.

\bibitem[\protect\citeauthoryear{Nakkiran \bgroup \em et al.\egroup
  }{2020}]{Nakkiran2020Deep}
Preetum Nakkiran, Gal Kaplun, Yamini Bansal, Tristan Yang, Boaz Barak, and Ilya
  Sutskever.
\newblock Deep double descent: Where bigger models and more data hurt.
\newblock In {\em International Conference on Learning Representations}, 2020.

\bibitem[\protect\citeauthoryear{Northcutt \bgroup \em et al.\egroup
  }{2021}]{northcutt2021pervasive}
Curtis~G Northcutt, Anish Athalye, and Jonas Mueller.
\newblock Pervasive label errors in test sets destabilize machine learning
  benchmarks.
\newblock In {\em Thirty-fifth Conference on Neural Information Processing
  Systems Datasets and Benchmarks Track (Round 1)}, 2021.

\bibitem[\protect\citeauthoryear{Power \bgroup \em et al.\egroup
  }{2021}]{power2022grokking}
Alethea Power, Yuri Burda, Harri Edwards, Igor Babuschkin, and Vedant Misra.
\newblock Grokking: Generalization beyond overfitting on small algorithmic
  datasets.
\newblock In {\em ICLR MATH-AI Workshop}, 2021.

\bibitem[\protect\citeauthoryear{Stephenson and
  Lee}{2021}]{stephenson2021epochwise}
Cory Stephenson and Tyler Lee.
\newblock When and how epochwise double descent happens, 2021.

\bibitem[\protect\citeauthoryear{Stephenson \bgroup \em et al.\egroup
  }{2021}]{stephenson2021on}
Cory Stephenson, suchismita padhy, Abhinav Ganesh, Yue Hui, Hanlin Tang, and
  SueYeon Chung.
\newblock On the geometry of generalization and memorization in deep neural
  networks.
\newblock In {\em International Conference on Learning Representations}, 2021.

\bibitem[\protect\citeauthoryear{Valiant}{1984}]{valiant1984theory}
L.~G. Valiant.
\newblock A theory of the learnable.
\newblock {\em Commun. ACM}, 27(11):1134–1142, nov 1984.

\bibitem[\protect\citeauthoryear{Wilson and
  Izmailov}{2020}]{DBLP:journals/corr/abs-2002-08791}
{Andrew Gordon} Wilson and Pavel Izmailov.
\newblock Bayesian deep learning and a probabilistic perspective of
  generalization.
\newblock {\em Advances in Neural Information Processing Systems},
  2020-December, 2020.

\bibitem[\protect\citeauthoryear{Zhang \bgroup \em et al.\egroup
  }{2017}]{DBLP:conf/iclr/ZhangBHRV17}
Chiyuan Zhang, Samy Bengio, Moritz Hardt, Benjamin Recht, and Oriol Vinyals.
\newblock Understanding deep learning requires rethinking generalization.
\newblock In {\em 5th International Conference on Learning Representations,
  {ICLR} 2017, Conference Track Proceedings}, 2017.

\bibitem[\protect\citeauthoryear{Zhang \bgroup \em et al.\egroup
  }{2019}]{zhang2019layers}
Chiyuan Zhang, Samy Bengio, and Yoram Singer.
\newblock Are all layers created equal?, 2019.

\end{thebibliography}
